\pdfoutput=1

\documentclass[11pt]{article}

\usepackage[]{acl}

\usepackage{times}
\usepackage{mathtools}
\usepackage{latexsym}
\usepackage{times}
\usepackage{url}
\usepackage{latexsym}
\usepackage[utf8]{inputenc}
\usepackage{amsmath}
\usepackage{amssymb}
\usepackage{times}
\usepackage{helvet}
\usepackage{courier}
\usepackage{url}
\usepackage{graphicx}
\usepackage{amsfonts}
\usepackage{xspace}
\usepackage{verbatim}
\usepackage{CJKutf8}
\usepackage{multirow}
\usepackage{tabularx}

\usepackage{array}
\usepackage{adjustbox}
\usepackage{booktabs}
\usepackage{amsthm}
\usepackage{subcaption}
\usepackage{algorithm}
\usepackage{algorithmic}

\usepackage{color}
\usepackage{colortbl}
\usepackage{xcolor}
\usepackage{relsize}
\usepackage{cleveref}
\usepackage{bbm}
\usepackage{xcolor}
\usepackage{soul}
\usepackage{xspace}

\definecolor{RED}{RGB}{255,0,0}
\definecolor{GREEN}{RGB}{0,172,78}

\newcommand{\hlc}[2][yellow]{{%
		\colorlet{foo}{#1}%
		\sethlcolor{foo}\hl{#2}}%
}

\usepackage[T1]{fontenc}

\usepackage[utf8]{inputenc}
\newcommand{\approach}{\textsc{REGEX}}
\usepackage{microtype}
\newcommand\integratedgrads{\textit{IG}}
\newcommand{\synteq}{::=}
\theoremstyle{definition}

\title{Towards Faithful Explanations for Text Classification with Robustness Improvement and Explanation Guided Training}
\author{Dongfang Li$^1$, Shan He$^1$, Baotian Hu$^1$,  Qingcai Chen$^{1,2}$\\
$^1$Harbin Institute of Technology (Shenzhen), Shenzhen, China \\
$^2$Peng Cheng Laboratory, Shenzhen, China\\
\texttt{crazyofapple@gmail.com}}
\begin{document}
\maketitle
\begin{abstract}
Feature attribution methods highlight the important input tokens as explanations to model predictions, which have been widely applied to deep neural networks towards trustworthy AI. However, recent works show that explanations provided by these methods face challenges of being faithful and robust. In this paper, we propose a method with \textbf{R}obustness improvement and \textbf{E}xplanation \textbf{G}uided training towards more faithful \textbf{EX}planations (\textbf{\approach{}}) for text classification. First, we improve model robustness by input gradient regularization technique and virtual adversarial training. Secondly, we use salient ranking to mask noisy tokens and maximize the similarity between model attention and feature attribution, which can be seen as a self-training procedure without importing other external information. We conduct extensive experiments on six datasets with five attribution methods, and also evaluate the faithfulness in the out-of-domain setting. The results show that~\approach{} improves fidelity metrics of explanations in all settings and further achieves consistent gains based on two randomization tests. Moreover, we show that using highlight explanations produced by \approach{} to train select-then-predict models results in comparable task performance to the end-to-end method. 
\end{abstract}

\section{Introduction}
\begin{figure}[ht!]
\centering
\resizebox{\linewidth}{!}{
\includegraphics[width=0.8\textwidth]{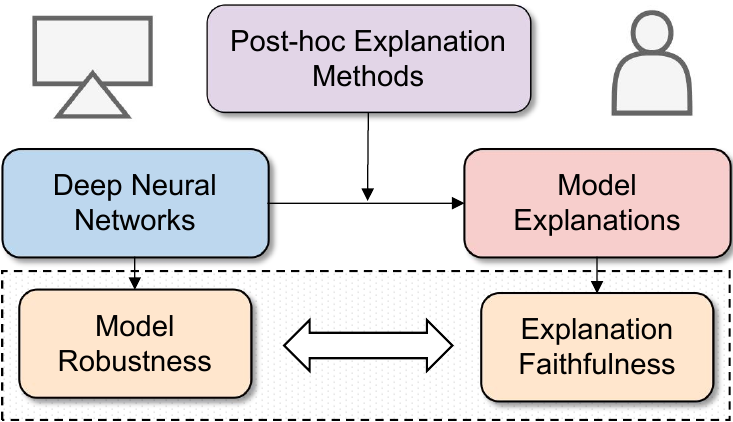}
}
\begin{tabular}{|ll|}
    \hline

     Model A: & It is \hlc[cyan!10]{a} \hlc[red!40]{warm} \hlc[red!60]{funny} \hlc[red!40]{engaging} \hlc[cyan!20]{film} . \\
     
     Model A': & It is \hlc[cyan!5]{a} \hlc[cyan!40]{warm} \hlc[cyan!20]{funny} \hlc[red!20]{engaging} \hlc[cyan!30]{film} . \\
    \hline
  \end{tabular}
\vspace{-2mm}
\caption{Visualization of \hlc[red]{positive} and \hlc[cyan]{negative} highlights produced by post-hoc explanation methods (e.g., feature attribution). However, these explanations suffer from unfaithfulness problems (e.g., same model framework A and A' with different attributions) and can be further fooled by adversarial manipulation without changing model output~\cite{Fragile} (see \S\ref{sec:explan_robust}).}
\vspace{-4mm}
\label{fig:schma}
\end{figure}
As the broad adoption of Pre-trained Language Models (PLMs) requires humans to trust their output, we need to understand the rationale behind the output and even ask questions regarding how the model comes to its decision~\cite{DBLP:journals/queue/Lipton18}. Recently, explanation methods for interpreting why a model makes certain decisions are proposed and become more crucial. For example, feature attribution methods assign scores to tokens and highlight the important ones as explanations~\cite{IG,FRESH,deyoung2019eraser}.

However, recent studies show that these explanations face challenges of being faithful and robust~\cite{DBLP:conf/nips/YehHSIR19,sinha-etal-2021-perturbing,ivankay2022fooling}, illustrated in Figure~\ref{fig:schma}. The \textit{faithfulness} means the explanation accurately represents the reasoning behind model predictions~\cite{Jacovi2020TowardsFI}. Though some works are proposed to use higher-order gradient information~\cite{smilkov2017smoothgrad}, by incorporating game-theoretic notions~\cite{DBLP:conf/iclr/HsiehYLRKKH21} and learning from priors~\cite{Enjoy_the_Salience}, how to improve the faithfulness of highlight explanations remains an open research problem. Besides, the explanation should be stable between functionally equivalent models trained from different initializations~\cite{DBLP:conf/acl/ZafarDSADK21}. Intuitively, the potential causes of these challenges could be (i)  the model is not robust and mostly leads to unfaithful and fragile explanations~\cite{AM,unifying} and (ii) those explanation methods themselves also lack robustness to imperceptible perturbations of the input~\cite{Fragile}; hence we need to develop better explanation methods.  In this paper, we focus on the former and argue that there are connections between model robustness and explainability; any progress in one part may represent progress in both.

To this end, we propose a method with \textbf{R}obustness improvement and \textbf{E}xplanation \textbf{G}uided training to improve the faithfulness of \textbf{EX}planations (\textbf{\approach{}}) while preserving the task performance for text classification. First, we apply the input gradient regularization technique and virtual adversarial training to improve model robustness. While previous works found that these mechanisms can improve the adversarial robustness and interpretability of deep neural networks~\cite{RossD18,unifying}, to the best of our knowledge, the faithfulness of model explanations by applying them has not been explored.  Secondly, our method leverages token attributions aggregated by the explanation method, which provides a local linear approximation of the model's behaviour~\cite{DBLP:journals/jmlr/BaehrensSHKHM10}. We mask input tokens with low feature attribution scores to generate perturbed text and then maximize the similarity between new attention and attribution scores. Furthermore, we minimize the Kullback–Leibler (KL) divergence between model attention of original input and attributions. The main idea is to allow attention distribution of the model to learn from input importance during training to reduce the effect of noisy information.

To verify the effectiveness of \approach{}, we consider a variety of classification tasks across six datasets with five attribution methods. Additionally, we conduct extensive empirical studies to examine the faithfulness of five feature attribution approaches in out-of-domain settings. The results show that \approach{} improves the faithfulness of the highlight explanations measured by sufficiency and comprehensiveness~\cite{deyoung2019eraser} in all settings while outperforming or performing comparably to the baseline, and further achieves consistent gains based on two randomization tests. Moreover, we show that using the explanations output from \approach{} to train select-then-predict models results in comparable task performance to the end-to-end method, where the former trains an independent classifier using only the rationales extracted by the pre-trained extractor~\cite{FRESH}. 
Considering neural network models may be the primary source of fragile explanations~\cite{ju-etal-2022-logic,tang2022identifying}, our work can be seen as a step towards understanding the connection between explainability and robustness -- the desiderata in trustworthy AI.
The main contributions of this paper can be summarized as:
\begin{itemize}
    \item We explore how to improve the faithfulness of highlight explanations generated by feature attributions in text classification tasks. 
    \item We propose an explanation guided training mechanism towards faithful attributions, which encourages the model to learn from input importance during training to reduce the effect of noisy tokens.
    \item We \textit{empirically} demonstrate that REGEX models generate more faithful explanations by \textit{extensive} experiments on 6 datasets and 5 methods, which suggests that the faithfulness of highlight explanations may be improved by considering model robustness.\footnote{We will publicly release the code, pre-trained models and all experimental setups.}
\end{itemize}

\begin{figure*}[hpt]
\centering
%
\includegraphics[width=0.7\linewidth]{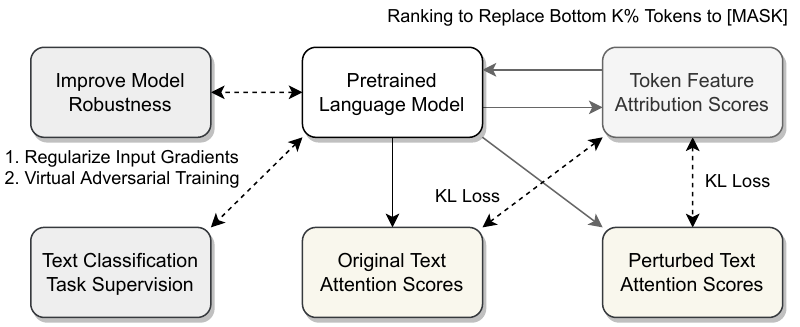}
\vspace{-2mm}
\caption{The overall framework of proposed \approach{} method. \approach{} consists of two components for robustness improvement and explanations guided training respectively. For latter, we iteratively mask input tokens with low attribution scores and then minimize the KL divergence between attention of masked input and feature attributions. }
\vspace{-4mm}
\label{fig:model}
\end{figure*}
\section{Related Work}
\paragraph{Model Robustness and Explainability} 
As it has recently been shown that deep neural networks are vulnerable to adversarial attacks even with PLMs, several works are proposed to ensure that AI systems are trustworthy and reliable, which include quantifying the vulnerability and designing new attacks and better defense technologies~\cite{DBLP:conf/acl/HendrycksLWDKS20,DBLP:journals/corr/abs-2112-08313}. 
However, as the debug tools for black-box models, explanation methods also lack robustness to imperceptible and targeted perturbations of the input~\cite{DBLP:conf/nips/HeoJM19,DBLP:journals/corr/abs-1910-02065,DBLP:conf/acl/MeisterLAC20,DBLP:conf/iclr/HsiehYLRKKH21}. While significantly different explanations are provided for similar models~\cite{DBLP:conf/acl/ZafarDSADK21}, how to elicit more reliable explanations is a promising direction towards interpretation robustness. Different from~\citet{DBLP:conf/acl/CamburuSMLB20} that addresses the inconsistent phenomenon of explanations, we investigate the connection between model robustness and faithfulness of the explanations.
\paragraph{Explanation Faithfulness} 
The faithfulness of explanations is important for NLP tasks, especially when humans refer to model decisions~\cite{DBLP:journals/corr/abs-1711-00867,DBLP:journals/corr/abs-1809-10804}.~\citet{Jacovi2020TowardsFI} first propose to evaluate the faithfulness of Natural Language Processing (NLP) methods by separating the two definitions between faithfulness and plausibility and provide guidelines on how evaluation of explanations methods should be conducted. Recently, some works have focused on faithfulness measurements of NLP model explanations and improve the faithfulness of specific explanations~\cite{DBLP:conf/emnlp/WiegreffeMS21,DBLP:journals/corr/abs-2104-08782,DBLP:conf/acl/ChrysostomouA20,DBLP:journals/corr/abs-2111-07367}. 
Among them,~\citet{DBLP:conf/naacl/DingK21} propose two specific consistency tests intending to measure if the post-hoc explanations remain consistent with similar models.

\paragraph{Incorporate Explanations into Learning}

While most previous explanation methods have been developed for explaining deep neural networks, some works explore the potential to leverage these explanations to help build better models~\cite{liuAcl19,DBLP:conf/icml/RiegerSMY20,DBLP:conf/emnlp/JayaramA21,DBLP:conf/emnlp/JuZT0CZLZ21,DBLP:journals/corr/abs-2109-08259,DBLP:conf/emnlp/HanT21,DBLP:journals/corr/abs-2111-14338,Enjoy_the_Salience,stacey2021supervising,DBLP:journals/corr/abs-2110-07586}. For example,~\citet{DBLP:journals/corr/abs-2102-02201} propose a framework to understand the role of explanations in learning, and find that explanations are suitably used in a retrieval-based modeling approach. Similarly, ~\citet{adebayo2022post}  investigate whether post-hoc explanations effectively detect model reliance on spurious training signals, but the answer seems to be negative. While effectively incorporating explanations remains an open problem, we focus on using model explanations in a self-training way to improve its faithfulness.

\section{Method}
\subsection{Problem Formulation}

First, we consider the setting of multi-label text classification problem with $n$ input examples $\{(\boldsymbol{x}_i,y_i)\}_{i=1}^n$. The input space embedded into vectors is $\mathcal{\boldsymbol{x}} \subseteq \mathbb{R}^{l\times d}$ and the output space is $\mathcal{Y}$. A neural classifier is $f_\theta: \mathcal{X} \rightarrow \mathcal{Y}$ where $f_\theta(\boldsymbol{x})$ parameterized by $\theta$ which denotes the output class for one example $\boldsymbol{x} = \left(x_{1},~\cdots, x_{l}\right) \in \mathcal{X}$, where $l$ represents the length of the sequence. The optimization of the network is to minimize the cross-entropy loss $\mathcal{L}$ over the training set as follows:
\begin{equation}
\label{classify_loss}
    \mathcal{L}_{classify}= -\sum_{i=1}^n \log p_\theta(y_i | \boldsymbol{x}_i).
\end{equation}
Then, given an input $\boldsymbol{x}_i = (x_1,~\cdots, x_l)$ and its particular prediction $f_\theta(\boldsymbol{x}_i) = y_i$, the goal of feature attribution is to assign each token with a normalized score that then can be used to extract a compact set of relevant sub-sequences  with respect to the prediction. Formally, an attribution of the prediction at input $\boldsymbol{x}_i$ is a vector $\boldsymbol{
a}_i = (a_{i1},~\cdots, a_{il})$ and $a_{ij}$ is defined as the attribution of $x_{ij}$.
After that, we denote the set of extracted tokens (i.e., highlight explanations or rationales) provided by taking top-$k$ values from $\boldsymbol{x}_i$ as $\boldsymbol{r}_i$, and use $\boldsymbol{\overline{r}}_i = \boldsymbol{x}_i \setminus \boldsymbol{r}_i$, as the complementary set of $\boldsymbol{r}_i$ to denote the set of irrelevant tokens. 

\subsection{Robustness Improvement}
Adversarial attacks are inputs that are intentionally constructed to mislead neural networks ~\cite{szegedy2013intriguing,goodfellow2014explaining}. 
Given the $f_{\theta}$ and an input $\boldsymbol{x} \in \mathcal{X}$ with the label $y \in \mathcal{Y}$, an adversarial example $\boldsymbol{x}_{adv}$ satisfies
   \vspace{-2mm} 
   \begin{eqnarray}
    \label{eq:adv}
    &\boldsymbol{x}_{adv}=\boldsymbol{x}+\mathbf{\epsilon},  f(\boldsymbol{x})=y \wedge f(\boldsymbol{x}_{adv}) \neq y 
    \end{eqnarray}
where $\epsilon$ is the worst-case perturbation. 
Several defense methods have been proposed to increase the robustness of deep neural networks to adversarial attacks.  We adopt two popular methods: \textit{virtual adversarial training}~\cite{Miyato2015DistributionalSW} which leverages a regularization loss to promote the smoothness of the model distribution, and \textit{input gradient regularization}~\cite{RossD18} which regularizes the gradient of the cross-entropy loss. Note that the methods used to improve the robustness are not limited to these techniques.

As shown in Figure~\ref{fig:model}, we aim to improve the robustness of deep neural networks intrinsically. Instead of adopting adversarial training objective, we follow~\citet{jiang2019smart} to regularize the standard objective using virtual adversarial training \cite{miyato2018virtual}:
\begin{equation}
  \mathcal{L}_{at}(\boldsymbol{x}, y, \theta) = \max_{\delta} l(f(\boldsymbol{x}+\delta; \theta), f(\boldsymbol{x}; \theta)).
\end{equation} The goal of this approach is the enhancement of label smoothness in the embedding neighborhood. Specially, we run additional projected gradient steps to find the perturbation $\delta$ with violation of local smoothness to maximize the adversarial loss.
On the other hand, input gradient regularization trains neural networks by minimizing not just the ``energy'' of the network but the rate of change of that energy with respect to the input features~\cite{DBLP:journals/tnn/DruckerL92}.
The goal of this approach is to ensure that if any input changes slightly, the KL divergence between the predictions and the labels will not change significantly. Formally, it takes the original loss term and penalizes the $\ell_{2}$ norm of its gradient and parameters:
\begin{equation}
    \mathcal{L}_{gr}(\boldsymbol{x}, y, \theta) = \lVert{\frac{\partial}{\partial \boldsymbol{x}}\mathcal{L}(\boldsymbol{x}, y, \theta)}\rVert_2 + \lVert{ \theta}\rVert_2.
\end{equation} It can also be interpreted as applying a particular projection to the Jacobian of the logits and regularizing it~\cite{RossD18}.

\subsection{Explanation Guided Training}
\label{section:egt}
If post-hoc explanations faithfully quantify the model predictions, the irrelevant tokens should have low feature attribution scores~\cite{DBLP:journals/corr/abs-2111-14338}. Based on this intuition, we leverage the existing explanations to guide the model for reducing feature attribution scores of irrelevant tokens without sacrificing the model performance. Concretely, we propose the Explanation Guided Training (EGT) mechanism. Instead of using the saliency method (i.e., gradient of the target class with respect to the input)~\cite{DBLP:journals/corr/SimonyanVZ13}, we apply the \textit{Integrated Gradients} (IG) method~\cite{IG} that is more faithful via axiomatic proofs to calculate the token importance. We do not assume the IG is totally faithful, and we also experiment with other attribution methods in \S\ref{sec:ablation}. It integrates the gradient along the path from an uninformative baseline to the original input. 
This baseline input is used to make a high-entropy prediction that represents uncertainty. As it takes a straight path between baseline and input, it requires computing gradients several times. The motivation for using path integral rather than vanilla gradient is that the gradient might have been saturated around the input while the former can alleviate this problem. 
Formally, given an input $\boldsymbol{x}$ and baseline $\boldsymbol{x'}$, the integrated gradient along the $i^{th}$ dimension is defined as follows:
\begin{equation}
\begin{aligned}
\small
\integratedgrads_{i}(\boldsymbol{x}) \synteq (x_i-{x'}_i) \int_{\alpha=0}^{1} \tfrac{\partial f_\theta(\boldsymbol{x'} + \alpha\times(\boldsymbol{x}-\boldsymbol{x'}))}{\partial x_i  }~d\alpha,
\end{aligned}
\end{equation}
where
$\tfrac{\partial f_\theta(\boldsymbol{x})}{\partial x_i}$ represents the gradient of $f$ along the $i^{th}$ dimension at $\boldsymbol{x}$ which is the concatenated embedding of the input sequence, and the attribution of each token is the sum of the attributions of its embedding. Note that we attribute the output of the model with ground-truth labels during training. We also test other feature attribution methods in \S\ref{sec:ablation}.

After calculating the token's importance score by $\ell_{2}$ aggregation over embedding dimensions, we sort tokens of $\boldsymbol{x}$ based on these scores and mask the bottom $K$\% words according to that sorting. We define the sorting function as $\mathbf{s}(\cdot)$ and the masking function as $\mathbf{m}(\cdot)$. For example, $\mathbf{s}_i(\boldsymbol{x})$ is the $i^{th}$ smallest element in $\boldsymbol{x}$, and $\mathbf{m}_k(\mathbf{s}(\boldsymbol{x}),\boldsymbol{x})$ replaces all $x_{i} \in \{\mathbf{s}_{i}\left(\boldsymbol{x}\right) \}_{e=0}^{ceil(1,Kl)}$ with a mask distribution, i.e., $\mathbf{m}_k(\mathbf{s}(\boldsymbol{x}),\boldsymbol{x})$ removes the $K$\% lowest features from $\boldsymbol{x}$ based on the order provided by $\mathbf{s}(\boldsymbol{x})$.
During training, we generate a new input $\widetilde{ \boldsymbol{x}}$ for each example $\boldsymbol{x}$ by masking the features with low attribution scores as follows:
\begin{align}
\widetilde{ \boldsymbol{x}}= \mathbf{m}_k(\mathbf{s}_{IG}(\boldsymbol{x}),\boldsymbol{x}).
\end{align}
$\widetilde{ \boldsymbol{x}}$ is then passed through the network which results in an attention scores $att(\widetilde{ \boldsymbol{x}})$. Following~\citet{FRESH}, the attention scores are taken as the mean self-attention weights induced from the first token index to all other indices. Then we maximize the similarity between $att\left(\boldsymbol{x}\right)$ and $att(\widetilde{ \boldsymbol{x}})$ to ensure that the model produces similar output probability distributions over labels for both masked and unmasked inputs. The optimization objective for the EGT is:
\begin{equation}
\small
\begin{aligned}
\mathcal{L}_{kl}(\boldsymbol{x}, y, \theta) =  D_{KL} \left(att(\boldsymbol{x}) ;\ IG(x) \right) + \\ 
 D_{KL} \left(att(\widetilde{\boldsymbol{x}}) ;\ IG(x) \right),
\end{aligned}
\end{equation}
where $D_{KL}$ is the KL divergence function between two distributions. The motivation behind two KL divergence terms is to encourage the model to focus on high salient words and ignore low salient words during training, and generate similar outputs for the original input $\boldsymbol{x}$ and masked input $\widetilde{\boldsymbol{x}}$, which can be seen as a special adversarial example. On the other hand, as the calculation of the mask input is batch-wise, the model should learn to assign low gradient values to irrelevant tokens for the predicted label in an iterative way. 

\subsection{Training}

We define the final weighted loss as follows,
\begin{equation}
\small
\mathcal{L} = \lambda_1\mathcal{L}_{classify} + \lambda_2\mathcal{L}_{gr}  + \lambda_3\mathcal{L}_{at}  + \lambda_4\mathcal{L}_{kl},
\label{loss_function}
\end{equation}
where $\lambda_1$, $\lambda_2$, $\lambda_3$ and $\lambda_4$ are hyper-parameters. Mixing these losses requires multiple forward and backward propagations (2.1x training time), but not increases inference time. And in this process we do not introduce external knowledge, only use salient ranking as self-training. At inference, we calculate the label probability and use different explanation methods in \S\ref{para:feature_attribution} to generate highlight explanations.

\subsection{Erasure-based Faithfulness Evaluation}
To evaluate post-hoc explanations, we adopt \textit{sufficiency} that measures the degree to which the highlight explanation is adequate for a model to make predictions, and \textit{comprehensiveness} that measures the influence of explanations to predictions~\cite{deyoung2019eraser}. These two metrics are usually used to evaluate faithfulness as it does not require re-training and the main idea is to estimate the effect of changing parts of inputs on model output. Let $p_\theta(y^j|\boldsymbol{x}_i)$ be the output probability of the $j$-th class for the $i$-th example, and rationale $\boldsymbol{r}_i$ extracted according to attribution scores. Formally, the sufficiency we used is as follows:
\begin{equation}
\small
\text{S}(\boldsymbol{x}_i, y^j, \boldsymbol{r}_i) = 1 - max(0, p_\theta(y^j|\boldsymbol{x}_i) - p_\theta(y^j|\boldsymbol{r}_i)),
\end{equation}
\vspace{-2mm}
\begin{equation}
\small
     \text{sufficiency}(\boldsymbol{x}_i, y^j, \boldsymbol{r}_i) = \frac{\text{S}(\boldsymbol{x}_i, y^j, \boldsymbol{r}_i) - \text{S}(\boldsymbol{x}_i, y^j,\boldsymbol{0})}{1 - \text{S}(\boldsymbol{x}_i, y^j, \boldsymbol{0})},
\end{equation} where higher sufficiency values are better as we normalize and reverse it between 0 and 1, and $\text{S}(\boldsymbol{x}_i, y^j, \boldsymbol{0})$ is the sufficiency of the input where no token is erased. Similarly, we define the comprehensiveness as follows:
\begin{equation}
\small
\text{C}(\boldsymbol{x}_i, y^j, \boldsymbol{r}_i) = max(0, p_\theta(y^j|\boldsymbol{x}_i) - p_\theta(y^j|\boldsymbol{\overline{r}}_i)),
\end{equation}
\begin{equation}
\small
\text{comprehensiveness}(\boldsymbol{x}_i, y^j, \boldsymbol{r}_i) = \frac{\text{C}(\boldsymbol{x}_i, y^j, \boldsymbol{r}_i)}{1 - \text{S}(\boldsymbol{x}_i, y^j, \boldsymbol{0})},
\end{equation} where higher comprehensiveness values are better.
As choosing the appropriate rationale length is dataset dependent, we use the Area Over the Perturbation Curve (AOPC) metrics for sufficiency and comprehensiveness. It defines bins of tokens to be erased and calculates the average measures across bins. Here, we keep the top 1\%, 5\%, 10\%, 20\%, 50\% tokens into bins in the order of decreasing attribution scores.

\section{Experiments}
We conduct the experiments in six datasets under the in-domain/out-of-the-domain settings: SST~\cite{SST}, IMDB~\cite{IMDB}, Yelp~\cite{Yelp}, and AmazDigiMu/AmazPantry/AmazInstr~\cite{amazon} (See details in Appendix~\ref{sec:dataset:sta}). The baseline is a text classification model fine-tuned on the training set while the same pre-trained language model is applied to \approach{}. In other words, the baseline is optimized by Eqn.~\ref{classify_loss} without robustness improvement and explanation guided training mechanisms.

\subsection{Post-hoc Explanation Methods \label{para:feature_attribution}} We consider five feature attribution methods and a random attribution method:

\paragraph{\textbf{Random (\textsc{Rand})}}~\cite{previous_work}: Token importance is assigned at random.
\paragraph{\textbf{Attention ($\boldsymbol{\alpha}$)}}~\cite{FRESH}: Normalized attention scores are used to calculate token importance.
\paragraph{\textbf{Scaled Attention ($\boldsymbol{\alpha} \nabla \boldsymbol{\alpha}$)}}~\cite{serrano-smith-2019-attention}:Normalized attention scores $\alpha_i$ scaled by the corresponding gradients $\nabla \alpha_i= \frac{\partial \hat{y}}{\partial \alpha_i}$.
\paragraph{\textbf{InputXGrad ($\mathbf{x} \nabla \mathbf{x}$)}}~\cite{DBLP:journals/corr/ShrikumarGSK16,DBLP:journals/corr/KindermansSMD16}: The input attribution importance is generated by multiplying the gradient $\nabla x_i = \frac{\partial \hat{y}}{\partial x_i}$  with the input. 
\paragraph{\textbf{\textbf{Integrated Gradients ($\mathbf{IG}$)}}}~\cite{IG}: See \S\ref{section:egt} for details.

\paragraph{\textbf{\textbf{DeepLift}}}~\cite{deeplift}: The difference between each neuron activation and a reference vector is used to rank words.

\begin{table*}[!t]
    \centering
    \small
    \setlength{\tabcolsep}{0.8pt}\resizebox{\linewidth}{!}{
    \begin{tabular}{cc|cccccc|cccccc}
    \hline
         \multirow{2}{*}{\textbf{Train}} & \multirow{2}{*}{\textbf{Test}}  &  \multicolumn{6}{c|}{\textbf{Normalized Sufficiency ($\uparrow$}) } & \multicolumn{6}{c}{\textbf{Normalized Comprehensiveness ($\uparrow$}) } \\ 
          &  & \textsc{Rand} & $\alpha\nabla\alpha$ & $\alpha$ &  DeepLift &       $x\nabla x $ &        IG  & \textsc{Rand} & $\alpha\nabla\alpha$ & $\alpha$ &  DeepLift &       $x\nabla x $ &        IG  \\\hline
        \multirow{3}{*}{SST} &   SST   &.30(.38) &	.68(.51) & 	.48(.42) & 	.71(.42)& 	.49(.40) &	.49(.41) &	.22(.19) &	.56(.39) &	.41(.22) &	.52(.25)	& .43(.26) &	.43(.26) \\ 
              &        IMDB   & .25(.31)	& .54(.53) & 	.45(.39) &	.46(.32) &	.40(.31) &	.40(.32)	& .19(.23) &	.75(.54) &	.66(.34)& .61(.27) &	.58(.27)	& .58(.28)  \\
              &        Yelp   & .24(.32)	& .51(.56)	& .38(.40)	& .45(.35)	& .35(.33)	& .36(.34)&	.22(.21)& .70(.48)	& .57(.28)	&.59(.24)	& .48(.24) &	.47(.25)  \\\hline 
        \multirow{3}{*}{IMDB}  & IMDB   & .34(.32) &	.82(.55) &	.51(.46) & 	.80(.36) &	.54(.36)&	.53(.36)	& .17(.16) &	.71(.48) & 	.39(.31) &	.62(.25) & 	.31(.23) & 	.32(.24)  \\
             &         SST   & .30(.24) & 	.72(.35) & 	.42(.28)& .68(.28)	& .46(.27) &	.45(.27) & 	.21(.27)	& .59(.46)	& .28(.32)	& .51(.33)	& .32(.33)	& .33(.33)  \\
             &        Yelp   & .32(.35)	& .81(.48) & 	.53(.41) & 	.79(.36) & 	.48(.36)	& .47(.36)	& .20(.21)	& .71(.45)	& .42(.32) &	.64(.26) & 	.33(.26) &	.34(.26)  \\   \hline
         \multirow{3}{*}{Yelp} &  Yelp   & .35(.23)	& .82(.32) & 	.59(.31)	& .82(.29)	& .53(.24) & 	.53(.25) &	.10(.12)	& .64(.20)	& .39(.14)	& .63(.16)	& .24(.15) &	.23(.16) \\
             &         SST  & .33(.41) 	& .76(.45)	& .49(.43) &	.75(.44) &	.60(.41)	& .60(.41)&	.16(.17) &	.57(.24) &	.31(.18)	& .55(.21)	& .40(.22) &	.40(.22)  \\
             &        IMDB   &  .38(.18)	& .83(.34) & 	.59(.32) &	.82(.25) & 	.61(.22) &	.61(.22) &	.13(.19)& .74(.34) &	.43(.29) &	.70(.23) &	.31(.23) &	.31(.24) \\ \hline 
        \multirow{3}{*}{AmazDigiMu} &  AmazDigiMu   &  .50(.34)	& .73(.56) &	.55(.34) &	.66(.31) &	.60(.41) &	.62(.39)	& .18(.13)	& .60(.32)	& .12(.14)	& .21(.10) &	.26(.16) & 	.24(.17)  \\
       &   AmazInstr  & .60(.29) &	.75(.54)	& .67(.32)	& .67(.31) &	.66(.33) &	.68(.32) 	& .16(.19)	& .62(.47) &	.18(.23) &	.15(.19) & 	.24(.22) &	.23(.23)  \\
       &  AmazPantry  & .53(.33)&	.70(.55)	& .60(.33) & 	.64(.31) &	.60(.37) &	.62(.36)	& .19(.21)	& .61(.46)	& .13(.22) &	.18(.17) &	.24(.23) &	.22(.25) \\   \hline
        \multirow{3}{*}{AmazPantry} & AmazPantry   & .55(.25)	& .79(.46) & 	.56(.36) & 	.82(.19) & 	.54(.28)	& .52(.27)	& .15(.20) & 	.50(.42) &	.14(.31) &	.52(.15) &	.16(.25) &	.17(.25)  \\
       &  AmazDigiMu   & .54(.24) &	.78(.47) &	.56(.37) &	.82(.19) &	.52(.27) & .50(.26) & 	.14(.19) &	.50(.41) &	.16(.32)&	.52(.15)	& .14(.23) &	.15(.24)  \\
       &   AmazInstr   & .55(.17)	& .81(.42)	& .53(.30) & 	.82(.15) &	.51(.20) &	.50(.20) &	.14(.24) &	.60(.52) &	.13(.40) &	.60(.23) &	.15(.30)	& .16(.30)  \\ \hline
        \multirow{3}{*}{AmazInstr} &  AmazInstr  &  .52(.16) &	.82(.34) &	.58(.18) &	.82(.21) &	.59(.18) &	.58(.17)	& .16(.26) &	.58(.52) &	.22(.26)	& .56(.29)	& .18(.28) & 	.19(.29) \\
        &  AmazDigiMu   & .56(.21)	& .82(.38) & 	.58(.21) &	.82(.22) &	.60(.24) &	.59(.22)	& .12(.23)	& .48(.46) &	.16(.20) &	.46(.22) &	.15(.24) &	.15(.25)  \\
        &  AmazPantry  &  .56(.22) &	.81(.39)	& .58(.21) & .81(.23)	& .59(.24) & 	.58(.23) &	.13(.27) &	.50(.51) 	& .16(.22) &	.47(.25)	& .16(.27) & 	.17(.29)  \\ \hline
        
    \end{tabular}}
     \vspace{-2mm}
    \caption{Normalized sufficiency and comprehensiveness in the in- and out-of-domain settings for five feature attribution approaches and a random attribution. \approach{} vs. baseline (shown in brackets). For example, a value of .30 (.38) represents the result of Normalized Sufficiency on the SST test set with the RAND method, .30 means the score of our method, and .38 means the baseline.   
    }
    \vspace{-2mm}
    \label{tab:faithfulness_feature_scoring_comp-aopc}
\end{table*}

\begin{table*}[!t]
    \centering
    \small
    \begin{tabular}{l|ll|ccccc} \hline
\textbf{ Train} &   \textbf{Test}& \textbf{Full-text F1} &  \textbf{$\boldsymbol \alpha \nabla \alpha$} &   \textbf{$\boldsymbol \alpha$} &    \textbf{DeepLift} &   \textbf{$\boldsymbol x \nabla x$} &          \textbf{IG} \\ \hline
         \multirow{3}{*}{SST (20\%)} &         SST &   89.7(90.1)  &          \bf{88.9}(87.7)  &         83.0(81.1)   &          87.3(84.4)  &  77.8(76.3)  &        77.8(76.8)  \\
                             &        IMDB &   83.4(84.3)  &         \bf{86.3}(81.8)  &      65.3(52.6)  &         81.1(64.0)  &  53.2(55.0)  &         53.2(56.3)  \\
                             &        Yelp &   87.8(87.9)  & \bf{90.2}(88.1)  &         76.5(72.6)  &         80.4(75.4)  &  64.4(59.6)  &  64.4(63.9)  \\\hline 
      \multirow{3}{*}{IMDB (2\%)} &        IMDB &   91.3(91.1)  &          \bf{88.9}(87.9)  &         79.2(80.4)  &          87.6(87.2)  &  59.1(59.8)  &         59.1(59.7)  \\
                             &         SST &   88.0(85.8)  &          \bf{80.6}(80.9)  &          71.8(71.8)  &      72.9(70.1)  &  65.7(69.6)  &         65.7(70.7)  \\
                             &        Yelp &   90.3(91.0)  &          \bf{90.4}(87.8)  &          72.7(82.0)  &          86.5(79.4)  &  70.5(69.0)  &  70.5(69.1)  \\\hline 
      \multirow{3}{*}{Yelp (10\%)} &        Yelp &   96.1(96.9)  &         96.3(94.0)  &          87.1(90.4)  &         \bf{97.1}(93.6)  &  71.2(70.5)  &         71.2(71.9)  \\
                             &         SST &   85.3(86.8)  &          \bf{82.0}(59.3)  &          58.1(69.8)  &          69.9(67.2)  & 67.6(67.7)  &         67.6(69.3)  \\
                             &        IMDB &   86.2(88.6)  &         \bf{86.7}(78.0)  &          51.5(64.5)  &          79.1(66.6)  &  48.0(53.0)  &  48.0(55.8)   \\\hline
                           
 \multirow{3}{*}{AmazDigiMu (20\%)} &  AmazDigiMu &   72.4(70.6)   &         \bf{67.9}(66.1)  &      62.5(63.4)  &  67.5(65.8) &  48.3(51.9)  &         48.3(65.8)  \\
                             &   AmazInstr &   60.3(61.2) &  \bf{60.9}(58.0) &  50.0(57.2) &  \bf{60.9}(57.4) &  39.0(46.0)  &         39.0(57.2)  \\
                             &  AmazPantry &   61.0(64.6)  &          \bf{60.1}(59.1)  &          46.3(56.5)  &          59.0(56.5)  &  38.8(44.8)  &  38.8(44.8)   \\\hline
 \multirow{3}{*}{AmazPantry (20\%)} &  AmazPantry &   71.3(70.2)  &         67.8(67.3)  &          59.6(62.6)  &          \bf{68.0}(67.2)  &  50.3(48.6)  &         50.3(48.7)  \\
                             &  AmazDigiMu &   60.1(59.5)  &  \bf{58.5}(57.7) &         51.5(54.6)  &         58.4(56.2)  &  42.7(41.2)  &         42.7(57.7) \\
                             &   AmazInstr &   65.7(64.5)  &  64.9(63.8) &          54.9(58.0)  &  \bf{65.5}(63.6) &  43.3(40.1)  &  43.3(40.3)  \\\hline 
  \multirow{3}{*}{AmazInstr (20\%)} &   AmazInstr &   72.9(71.5)  &          69.5(69.8)  &          63.1(62.1)  &        \bf{70.7}(69.7)  &  47.5(45.6)  &        47.5(48.6)  \\
                             &  AmazDigiMu &   60.7(61.3)  &  58.6(60.0) &      51.6(53.2)  &        \bf{58.9}(57.8)  &  43.7(43.8)  &         43.7(60.0) \\
                             &  AmazPantry &   67.9(68.2)  &         65.0(64.5)  &      55.8(56.3)  &          \bf{65.6}(63.1)  &  45.2(44.6)  &  45.2(47.6)   \\ \hline
\end{tabular}
  \caption{Average macro F1 results of Full-text and FRESH models with a prescribed rationale length. \approach{} vs. baseline (shown in brackets, averaged across 5 seeds). The reference performance (Full-text F1) is from the BERT-base model fine-tuned on the full text. Full results are in Appendix~\ref{sec:fullresults}. The bold numbers represent the results of the best FRESH model trained with rationales from REGEX model among five attribution methods.}
      \vspace{-2mm}
    \label{tab:F1_ece_select-then-predict-FRESH}
\end{table*}

\subsection{Post-hoc Explanations Faithfulness}
We conduct experiments on the faithfulness metrics (i.e., normalized sufficiency and normalized comprehensive) to compare the fidelity of different post-hoc explanation methods between the baseline and \approach{} models. We extract rationale $r$ from a model by selecting the top-$k$ most important tokens measured by these post-hoc explanation methods. Following~\citet{previous_work}, we also evaluate explanation faithfulness in out-of-domain settings without retraining models (i.e., zero-shot), and we follow their settings with six dataset pairs and a random attribution baseline. Especially the model has first trained on the source datasets, and then we evaluate its performance on the test set of the target datasets.

As shown in Table~\ref{tab:faithfulness_feature_scoring_comp-aopc}, \approach{} improves the explanation faithfulness with all five attribution methods by a large gap under most in- and out-of-domain settings. Among them, scaled attention and DeepLift perform better than others. For example, \approach{} surpasses the baseline in the sufficiency metric for the explanation extracted by DeepLift under all scenarios, while the comprehensiveness decreases when the model is trained in the AmazDigiMu dataset and tested in the AmazInstr dataset. It shows that \approach{} improves the fidelity of post-hoc explanations measured by sufficiency and comprehensiveness. Nevertheless, we observe a decrease in the comprehensiveness metrics for attention and IG on specific datasets. For example, considering the uncertainty of attention as an interpretable method~\cite{serrano-smith-2019-attention}, the fidelity metrics of attention attribution are inferior to the baseline on all three Amazon Reviews datasets. 

Overall, feature attribution approaches outperform random attributions of in- and out-of-domain settings in most cases. Moreover, results show that post-hoc explanation sufficiency and comprehensiveness are higher in in-domain test sets than in out-of-domain except for the Yelp dataset. On the other hand, as shown in Table~\ref{tab:F1_ece_select-then-predict-FRESH}, \approach{} improves performance or achieves similar task performance to the baseline on most out-of-domain datasets.


\subsection{Quantitative Evaluation by FRESH Method}
We further compare the average macro F1 of the FRESH classifier~\cite{FRESH} across five random seeds in the in- and out-of-domain settings. In short, FRESH is a select-then-predict framework, and the general process is that an extractor is first trained where the labels are induced by arbitrary feature importance scores over token inputs; then, an independent classifier is trained exclusively on rationales provided by the extractor which are assumed to be inherently faithful. Here, rationales extracted by the top-$k$ most important tokens are used as input to the classifier for training and test. 

As shown in Table~\ref{tab:F1_ece_select-then-predict-FRESH}, the best two methods are DeepLift and scaled attention, which achieve a similar performance as the original text input model in the in- and out-of-domain settings and is consistent with the faithfulness evaluation. For example, the FRESH classifier applying the DeepLift attribution method is higher than the baseline and outperforms the model with the full text input (97.1 vs. 96.9) on the Yelp dataset. 
It also illustrates that the performance depends on the choice of the feature attribution method.

\subsection{Explanation Robustness}
\label{sec:explan_robust}
\begin{table}[!t]
    \centering
    \resizebox{\linewidth}{!}{
    \begin{tabular}{l|c|c|c|c|c}
    \hline
        \textbf{Jaccard@25\%} & \texttt{Init\#1} & \texttt{Init\#2} & \texttt{Init\#3} & \texttt{Init\#4} & \texttt{\#Untrained}\\ \hline
        \texttt{Init\#1} & 1.0 & .44(.33) & .54(.34)	&.56(.34) & .28(.27) \\
        \texttt{Init\#2} & .44(.33) & 1.0 & .45(.44) & .41(.34) & .16(.17)\\ 
        \texttt{Init\#3} & .54(.34) & .45(.44) & 1.0 & .56(.36) & .22(.21)\\ 
         \texttt{Init\#4}  & .56(.34) & .41(.34) & .56(.36) & 1.0 & .12(.16)\\ 
        \texttt{\#Untrained} & .28(.27) & .16(.17) & .22(.21) & .12(.16) & 1.0 \\
        \hline
        
    \end{tabular}}
    \vspace{-2mm}
    \caption{Jaccard@25\% between the feature attributions (\approach{} vs. baseline, here we use scaled attention) for models with same architecture, with same data, and same learning schedule, except for randomly initial parameters. }
    \vspace{-4mm}
    \label{table:robustness_expls}
\end{table}

Following~\citet{DBLP:conf/acl/ZafarDSADK21}, we test \textit{implementation invariance} of feature attributions by Untrained Model Test (UIT) and Different Initialization Test (DIT). The UIT and DIT measure the consistency and calculate the Jaccard similarity between feature attributions generated by the post-hoc explanation method. We use Jaccard similarity for explanations extracted by top 25\% important tokens using the scaled attention method. If the two attributions are more similar, the Jaccard metric is higher. We compare the \approach{} and baseline by comparing two identical models trained from different initializations. The \texttt{\#Untrained} is a untrained model which randomly initialize the fully connected layers attached on top of the Transformer encoders.
As shown in Table~\ref{table:robustness_expls}, \approach{} achieves an improved performance than baseline. For example, \approach{} gets 0.56 while baseline gets 0.36 for \texttt{Init\#3} and \texttt{Init\#4}. As we expected, the similarity between explanations of the trained and untrained models is low, e.g., 0.12 between \texttt{Init\#4} and \texttt{\#Untrained}. It shows that improving faithfulness of explanations can strengthen interpretation robustness. However, the overall results between the two feature attributions are still low as 50\% of similarity comparisons are less than 0.5.

\section{Analysis}
\subsection{Ablation Study}
\label{sec:ablation}
\begin{figure}[t!]
\centering
    \subfloat[Sufficiency]{{\includegraphics[width=3.7cm]{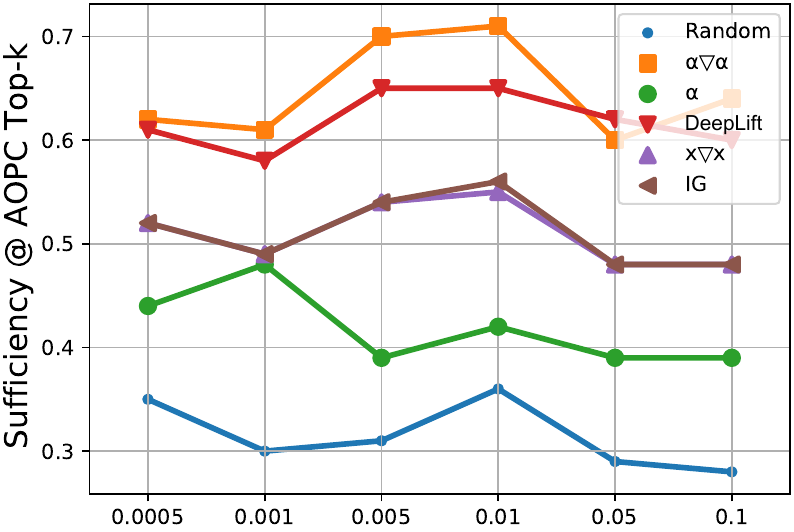}}}
    \subfloat[Comprehensiveness]{{\includegraphics[width=3.7cm]{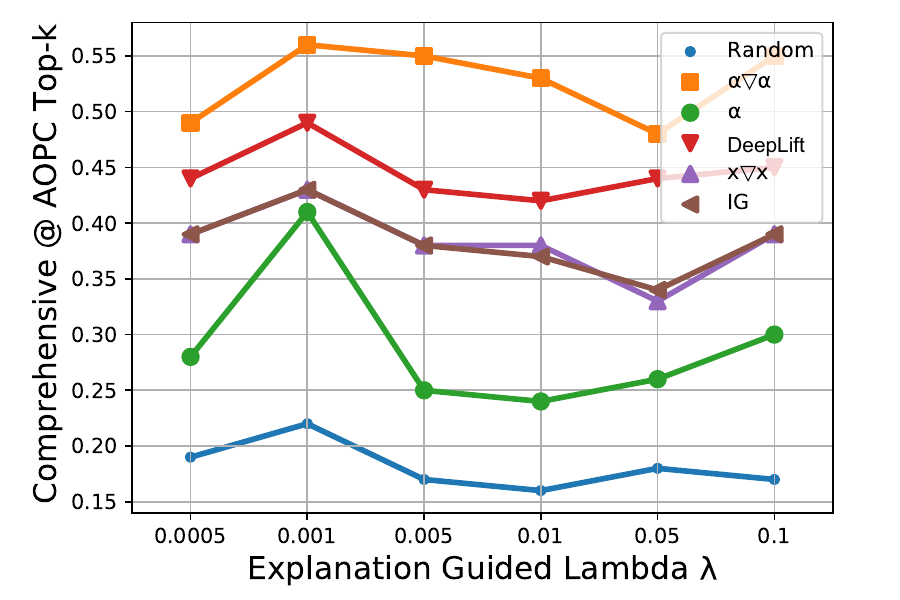}}}
\caption{Comparisons between different explanation guided training $\lambda_4$ on the SST dataset.}
\label{fig:ablationL}
\end{figure}

\begin{figure}[t!]
\centering

    \subfloat[Sufficiency]{{\includegraphics[width=3.7cm]{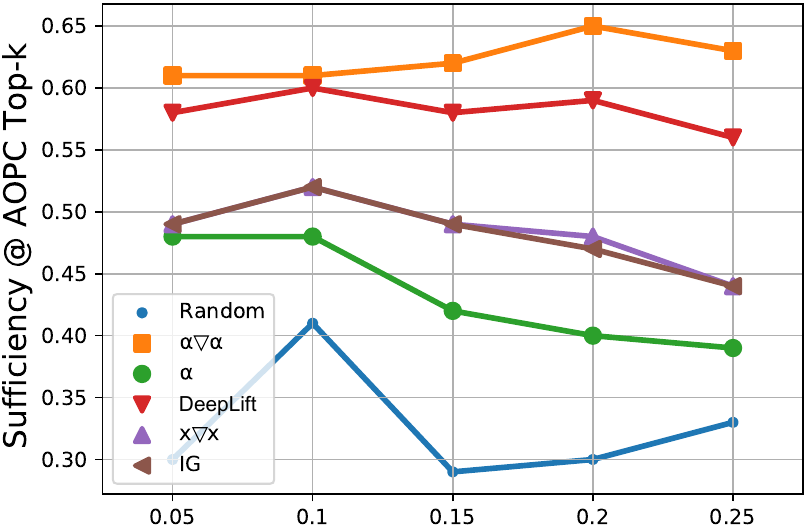}}}
    \subfloat[Comprehensiveness]{{\includegraphics[width=3.7cm]{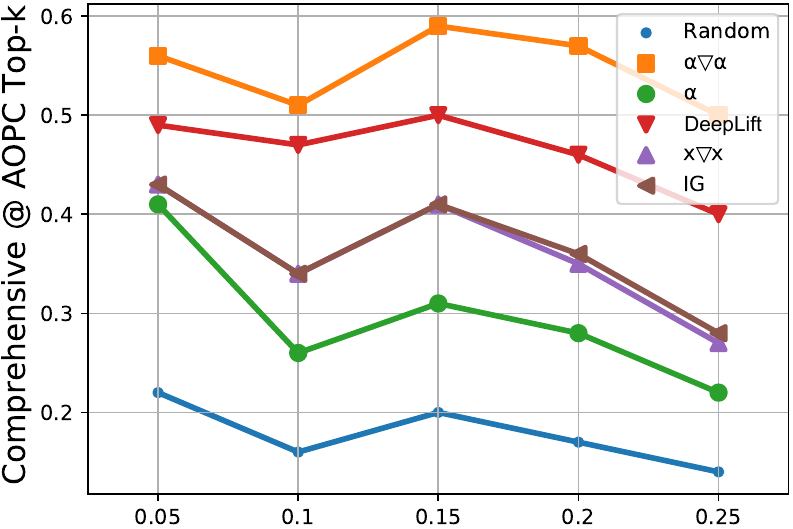}}}
\caption{Comparisons between different mask ratio $K$ on the SST dataset.}
\vspace{-2mm}
\label{fig:ablationK}
\end{figure}

We perform ablation studies to explore the effect of robustness improvement and explanation guided training for faithfulness evaluations shown in Table~\ref{tab:ablation} (all results in Table~\ref{tab:full_ablation}), and investigate the effect of different hyper-parameters on experimental results. We further compare the effect of the two aggregation methods (i.e., mean and $\ell_{2}$) during explanation guided training and the effect of using different feature attribution in \S\ref{section:egt} on the faithfulness of highlight explanations after training.
\paragraph{Robustness improvement is important for improving sufficiency and comprehensiveness.} Compared with \approach{} without explanation guided training, sufficiency and comprehensiveness of \approach{} without robustness improvement decrease more (0.14 vs. 0.02,  0.23 vs. 0.02, 0.29 vs. 0.07, 0.35 vs. 0.08).
\paragraph{The performance of the attention method varies more across different hyper-parameters.} In Figure~\ref{fig:ablationL}, we compare different $\lambda_4$ in Eqn.~\ref{loss_function} and observe that all methods achieve best sufficiency at 0.01 and best comprehensiveness at 0.001. In Figure~\ref{fig:ablationK}, we compare different mask ratios in \S\ref{section:egt} and find that the mask ratio between 0.15 and 0.2 is useful as larger values can bring noise.
\paragraph{The choice of aggregation method and feature attribution method in \S\ref{section:egt} has a large effect on the faithfulness evaluation.} We find that for most attribution methods, $\ell_{2}$ aggregation has higher fidelity performance. For example, Saliency with $\ell_{2}$ aggregation is better than Saliency with mean aggregation with more sufficiency improvement (0.70 vs. 0.55). Though there is no best method for explanation guided training, gradient-based methods (e.g., IG, 0.71) may be good choices in line with~\citet{inputxgrad}.

\begin{table}[t!]
    \centering
    \resizebox{\linewidth}{!}{
    \begin{tabular}{l|cccc}
    \hline
        \multirow{2}{*}{\textbf{Methods}} & \multicolumn{2}{c|}{\textbf{Suff.}} & \multicolumn{2}{c}{\textbf{Comp.}}    \\ \cline{2-3} \cline{4-5}
         & \textbf{$\boldsymbol \alpha \nabla \alpha$} & DeepLift & \textbf{$\boldsymbol \alpha \nabla \alpha$} & DeepLift  \\ \hline
        
        Saliency (Mean) & .52 & .48 & .48 & .42  \\ 
        InputXGrad (Mean) & .52 & .53 & .37 & .39  \\
         DeepLift (Mean) & .61 & .58 & .52 & .49  \\ 
        IG (Mean) & .47 & .45 & .49 & .51  \\ \hline
        Saliency ($\ell_{2}$) & \bf{.70} & .65 & .55 & .43  \\ 
        InputXGrad ($\ell_{2}$) & .58 & .54 & \bf{.58} & .49  \\ 
        DeepLift ($\ell_{2}$) & .69 & .68 & .53 & .47  \\ \hline
        \approach{} & .68 & \bf{.71} & .56 & \bf{.52} \\ 
        \quad w/o robustness improvement & .54 & .42 & .33 & .17  \\ 
        \quad w/o explanation guided training & .66 & .64 & .54 & .44  \\ \hline
    \end{tabular}}
    \caption{Ablation study with different aggregation methods and  feature attribution methods in \S\ref{section:egt}.}
    \label{tab:ablation}
\end{table}

\begin{table}[ht]
\includegraphics[width=\columnwidth]{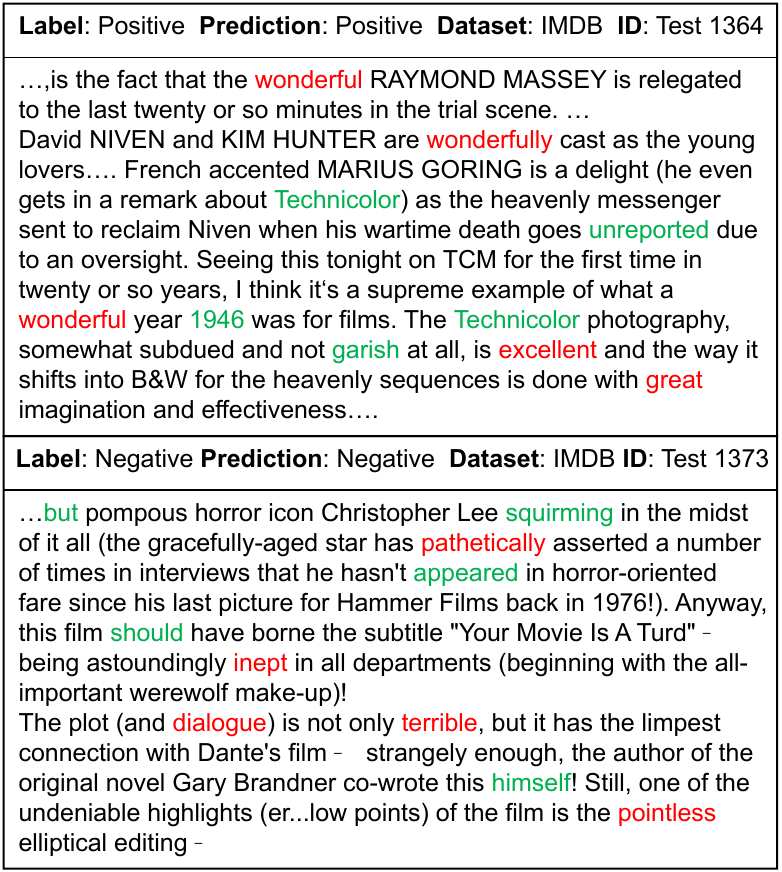}
\caption{We randomly pick two examples from test set of IMDB dataset, and highlight the Top-$k$ important tokens using DeepLift method (\textcolor{RED}{\approach{}} vs. \textcolor{GREEN}{Baseline}).}
\label{fig::case}
\vspace{-4mm}
\end{table}
\subsection{Qualitative Analysis}
Table~\ref{fig::case} presents two randomly-chosen examples of the test set of the IMDB dataset. For example, the top-$k$ important tokens returned by \approach{} are~\textit{wonderfully},~\textit{wonderful},~\textit{wonderful},~\textit{excellent} and~\textit{great} in the first example. We observe that these highlight explanations seem intuitive to humans and reasonably plausible. Though faithfulness and plausibility are not necessarily correlative~\cite{Jacovi2020TowardsFI}, we find that the highlights extracted by \approach{} contain more sentiment-related words, which should be helpful for review-based text classification.

\section{Conclusion}

We explore whether the fidelity of explanations can be further optimized and propose an explanation guided training mechanism. Extensive empirical studies are conducted on six datasets in both in- and out-of-domain settings. Results show that our method \approach{} improves both fidelity metrics and performance of select-then-predict models. The analysis of explanation robustness further shows that the consistency of explanations has been improved. The observation suggests that considering model robustness yields more faithful explanations. In the future, we would like to investigate more PLMs architectures and faithfulness metrics under the standard evaluation protocol. 

\section{Limitations}
Possible limitations include the limited PLM architecture/size (although we include additional results with RoBERTa in the Appendix~\ref{sec:roberta}) and faithfulness evaluation metrics are not necessarily comprehensive. And we only focus on text classification tasks. As a result, we do not investigate other language classification (e.g., natural language inference and question answering) and text generation tasks. If we can intrinsically know or derive the golden faithful explanations~\cite{DBLP:journals/corr/abs-2111-07367,2023arXiv230105062L}, the exploration of model robustness and explainability will be alternatively investigated for revealing the internal reasoning processes. And future work could include human study (e.g., evaluation about whether explanations help users choose the more robust of different models) and improve the robustness by more diverse ways (e.g., model distillation and data augmentation).

Our findings are also in line with~\citet{tang2022identifying} and Logic Trap 3~\cite{ju-etal-2022-logic}  which claims the model reasoning process is changed rather than the attribution method is unreliable. Different from this two works -- output probability perturbation or changing information flow, we view our results as complementary to their conclusion via sourcing the improvement of faithfulness. Although we show the link between robustness and faithfulness empirically, future work can strengthen the conclusions by discussion on a more conceptual and theoretical level. From a theoretical perspective, one possible reason is that the gradient of the model is more aligned with the normal direction to the close decision boundaries~\cite{DBLP:conf/icml/WangFD22}. In the future, we would like to analyze the relationship between robustness and explainability from~\textit{geometric dimension}.

Furthermore, we do not exhaustively experiment with all possible evaluation settings of interest even with the scale of our experiments. For example, saliency guided training methods~\cite{DBLP:journals/corr/abs-2111-14338} could have been used as another baseline. We hope this work inspires more future work that develops more effective strategies to make explanations reliable and investigate how our findings translate to large language models, such as GPT-3 model family\footnote{https://beta.openai.com/playground}, as with the emergent capabilities of these models, fidelity to their explanations or rationale will have societal impacts on accountability of NLP systems.


\bibliography{anthology,custom}
\bibliographystyle{acl_natbib}
\clearpage
\appendix
\label{sec:supplemental}
\begin{table}[t!]
    \centering 
    \small
    \begin{tabular}{l|c}\hline
         \textbf{Metric} & \textbf{Attack Results} \\ \hline

         Number of successful attacks: & 12(45) \\
         Original accuracy(\%):            & 93.0(96.0)  \\
         Accuracy under attack(\%):        & 90.0(84.8)   \\
         Attack success rate(\%):          & 3.23(11.71) \\
         Average perturbed word(\%):     &  39.06(27.02) \\
         Average num. words per input: & 244.73(244.73)   \\
         Avg num queries:              & 408.47(339.69)  \\ \hline
    \end{tabular}
    \caption{Attack results of~\approach{} and baseline by \textsc{CheckList} attack recipe.}
    \label{tab:model:robustness}
\end{table} 
\begin{table}[t]
    \centering
    \small
    \begin{tabular}{l|c}
    \hline
        \textbf{Methods} &  \textbf{Full-text F1} \\ \hline 
        
        Saliency (Mean) & 87.81$\pm$3.64  \\ 
        InputXGrad (Mean) &  91.21$\pm$0.23 \\
         DeepLift (Mean) & 87.99$\pm$0.48 \\ 
        IG (Mean) &  91.60$\pm$0.08 \\ \hline
        Saliency ($\ell_{2}$) & 83.52$\pm$1.29   \\ 
        InputXGrad ($\ell_{2}$) & 90.83$\pm$0.29\\ 
        DeepLift ($\ell_{2}$) &   87.62$\pm$0.53 \\ \hline
        \approach{} & 89.73$\pm$0.05 \\ 
        \quad w/o robustness improvement & 90.57$\pm$0.52  \\ 
        \quad w/o explanation guided training & 85.19$\pm$2.80  \\ \hline
    \end{tabular}
    \caption{Macro F1 and standard deviations with different aggregation methods and feature attribution methods in \S\ref{section:egt}.}
    \vspace{-4mm}
    \label{tab:ablationF1}
\end{table}
\section{Dataset}
\label{sec:dataset:sta}
We consider six datasets to evaluate explanations and the data statistics are as follows.

\paragraph{SST:} The Stanford Sentiment Treebank (SST) dataset~\cite{SST} includes review sentences (positive/negative) for analysis of the compositional effect of sentiment. The training set, development set, and test set consist of 6920, 872, and 1821 examples.

\paragraph{IMDB:} The IMDB dataset~\cite{IMDB} consists of 25k movies reviews from IMDB website labeled by sentiment (positive/negative). The training set, development set, and test set consist of 20k, 2.5k, and 2.5k examples.

\paragraph{Yelp:} The Yelp dataset~\cite{Yelp} includes highly polar movie reviews and is transformed to a binary classification task (positive/negative). The training set, development set, and test set consist of 476k, 84k, and 38k examples.

\paragraph{AmazDigiMu/AmazPantry/AmazInstr:} The amazon reviews dataset~\cite{amazon} is constructed by personalized justification from existing from Amazon review data. We choose the 3-class review and product metadata for three categories: Digital Music, Prime Pantry and Musical Instruments~\cite{previous_work}. These examples are then divided into three subsets: \textbf{AmazDigiMu} (122k/21k/25k examples), \textbf{AmazPantry} (99k/17k/20k examples) and \textbf{AmazInstr} (16k/29k/3k examples).

\section{Experiment Settings}
We use Spacy~\footnote{https://spacy.io/models/en} to pre-tokenize the sentence and apply the BERT-base model to encode text~\cite{bert}. 
We use AdamW optimizer with batch sizes of $8$, $16$, $32$, $64$  for model training. 
The initial learning rate is $1\times 10^{-5}$ for fine-tuning BERT parameters and $1\times 10^{-4}$ for the classification layer.
The maximum sequence length, the dropout rate, the gradient accumulation steps, the training epoch and the hidden size $d$ are set to $256$, $0.1$, 10\%, $10$, $768$ respectively. We clip the gradient norm within $1.0$. The learning parameters are selected based on the best performance on the development set. Our model is trained with NVIDIA Tesla A100 40GB GPUs (PyTorch \& Huggingface/Transformers~\footnote{https://github.com/huggingface/transformers} \& Captum~\footnote{https://captum.ai/}).  Following~\citet{jiang2019smart}, we set the perturbation size $\epsilon=1\times 10^{-5}$, the step size $\eta=1\times10^{-3}$, ascent iteration step $C=2$ and the variance of normal distribution $\sigma=1\times10^{-5}$. The weight parameters $\lambda_1$, $\lambda_2$, $\lambda_3$, $\lambda_4$ are set to  $1.0$, $0.01$, $0.5$, $0.01$ respectively. The mask ration $K$ is set to $0.15$. The number of steps used by the approximation method in IG is 50, and we use zero scalar corresponding to each input tensor as IG baselines. The parameters are selected based on the development set.
For the baseline and FRESH model, we use the same transformer-based models as mentioned previously to encode tokens and we choose rationale length by following~\citet{previous_work}. The model is trained for 10 epochs, and we keep the best models with respect to macro F1 scores on the development sets. 
\section{Text Classification to Attacks}
We conduct the behavioral testing with \textsc{CheckList}~\cite{checklist} and \texttt{TextAttack}~\cite{morris2020textattack} to attack \approach{} text classification models. We randomly choose 400 examples from IMDB test set as original attack examples, and the attack recipe greedily search adversarial examples to change the predicted label by contracting, extending, and substituting name entities in the sentence. The results are shown in the Table~\ref{tab:model:robustness} and the attack success rate which is used to evaluate the effectiveness of the attacks is 3.23\%.

\section{UIT and DIT with Larger Pre-trained Language Model}
\label{sec:roberta}
\begin{table}[!t]
    \centering
    \resizebox{\linewidth}{!}{
    \begin{tabular}{l|c|c|c|c|c}
    \hline
        \textbf{Jaccard@25\%} & \texttt{Init\#1} & \texttt{Init\#2} & \texttt{Init\#3} & \texttt{Init\#4} & \texttt{\#Untrained}\\ \hline
        \texttt{Init\#1} & 1.0 & .56(.40) & .60(.48)	&.61(.41) & .30(.31) \\
        \texttt{Init\#2} & .56(.40) & 1.0 & .50(.46) & .39(.36) & .20(.19)\\ 
        \texttt{Init\#3} & .60(.48) & .50(.46) & 1.0 & .55(.30) & .24(.25)\\ 
         \texttt{Init\#4}  & .61(.41) & .39(.36) & .55(.30) & 1.0 & .18(.18)\\ 
        \texttt{\#Untrained} & .30(.31) & .20(.19) & .24(.25) & .18(.18) & 1.0 \\
        \hline
        
    \end{tabular}}
    \caption{Jaccard@25\% between the feature attributions (\approach{} vs. baseline, scaled attention) based on RoBERTa~\cite{roberta} large model.}
    \label{table:robustness_expls_roberta}
\end{table}
To further verify the effect of model scale on the results, we conducted experiments on the robustness of explanations under the pre-trained language model RoBERTa~\cite{roberta}, including UIT and DIT. The experimental results are shown in the Table~\ref{table:robustness_expls_roberta}. We have two findings: (1) the size of the model has a certain positive effect on the stability of explanations, with the Jaccard similarity improved under REGEX and Baseline, although the improvement is not significant. (2) REGEX can still improve performance under larger pre-trained models which further strengths our findings.

\section{Full Results}
\label{sec:fullresults}
Table~\ref{tab:ablationF1} presents the Full-text F1 of variants in ablation study. Table~\ref{Fresh:full_results} lists the full results for FRESH (select-then-predict) models. Table~\ref{tab:full_ablation} lists the full results of ablation study.

From these results, we further found that \textbf{sufficiency of the extracted explanations when using one robustness training method (either virtual adversarial training or input gradient regularization) is inferior to the sufficiency when using no robustness training}. We speculate that there are several reasons: (1) the two mechanisms are related, i.e., removing one has a more significant impact than removing both simultaneously; (2) the results have variance despite the adoption of the AOPC metric, not to mention that the sufficiency metrics suffer from out-of-distribution challenges; (3) these ablation experiments are on models trained on SST and tested on SST; future works could perform a more detailed ablation analysis on other datasets (such as in out-of-domain settings). 
\begin{table*}[!t]
    \centering

    \begin{tabular}{llccccc} \hline
{ Train} &   {Test} &  {$\boldsymbol \alpha \nabla \alpha$} &   {$\boldsymbol \alpha$} &    {DeepLift} &   {$\boldsymbol x \nabla x$} &          {IG}                                  \\ \hline
           & SST        & {\color[HTML]{FF0000} {88.88$\pm0.7$}} & {\color[HTML]{FF0000} 83.00$\pm0.3$} & {\color[HTML]{FF0000} 87.31$\pm0.5$}          & {\color[HTML]{FF0000} 77.84$\pm0.5$} & {\color[HTML]{FF0000} 77.84$\pm0.5$} \\
SST        & IMDB       & {\color[HTML]{FF0000} {86.27$\pm0.2$}} & {\color[HTML]{FF0000} 65.32$\pm1.9$} & {\color[HTML]{FF0000} 81.18$\pm0.6$}          & 53.22$\pm0.6$                        & 53.22$\pm0.6$                        \\
           & Yelp       & {\color[HTML]{FF0000} {90.15$\pm0.1$}} & {\color[HTML]{FF0000} 76.45$\pm0.6$} & {\color[HTML]{FF0000} 80.35$\pm2.1$}          & {\color[HTML]{FF0000} 64.38$\pm0.5$} & {\color[HTML]{FF0000} 64.38$\pm0.5$} \\ \hline
           
           & IMDB       & {\color[HTML]{FF0000} {88.88$\pm0.3$}} & 79.16$\pm0.2$                        & {\color[HTML]{FF0000} 87.60$\pm0.2$}          & {\color[HTML]{FF0000} 59.14$\pm1.0$} & {\color[HTML]{FF0000} 59.14$\pm1.0$} \\
IMDB       & SST        & {80.60$\pm1.6$}                        & 71.75$\pm0.3$                        & {\color[HTML]{FF0000} 72.91$\pm0.6$}          & 65.68$\pm2.2$                        & 65.68$\pm2.2$                        \\
           & Yelp       & {\color[HTML]{FF0000} {90.37$\pm0.5$}} & 72.71$\pm1.0$                        & {\color[HTML]{FF0000} 86.51$\pm0.4$}          & {\color[HTML]{FF0000} 70.54$\pm0.9$} & {\color[HTML]{FF0000} 70.54$\pm0.9$} \\ \hline
          
           & Yelp       & {\color[HTML]{FF0000} 96.27$\pm0.1$}          & 87.13$\pm0.1$                        & {\color[HTML]{FF0000} {97.05$\pm0.0$}} & {\color[HTML]{FF0000} 71.22$\pm0.1$} & 71.22$\pm0.1$                        \\
Yelp       & SST        & {\color[HTML]{FF0000} {82.03$\pm0.5$}} & 58.13$\pm0.6$                        & {\color[HTML]{FF0000} 69.89$\pm0.4$}          & 67.58$\pm0.6$                        & 67.58$\pm0.6$                        \\
           & IMDB       & {\color[HTML]{FF0000} {83.68$\pm0.4$}} & 51.51$\pm0.4$                        & {\color[HTML]{FF0000} 79.10$\pm1.2$}          & 47.99$\pm1.8$                        & 47.99$\pm1.8$                        \\ \hline
         
           & AmazDigiMu & {\color[HTML]{FF0000} {67.87$\pm0.4$}} & 62.53$\pm0.9$                        & {\color[HTML]{FF0000} 67.52$\pm1.0$}          & 48.30$\pm2.2$                        & 48.30$\pm2.2$                        \\
AmazDigiMu & AmazInstr  & {\color[HTML]{FF0000} {60.95$\pm0.1$}} & 49.98$\pm0.8$                        & {\color[HTML]{FF0000} 60.92$\pm0.5$}          & 39.02$\pm0.2$                        & 39.02$\pm0.2$                        \\
           & AmazPantry & {\color[HTML]{FF0000} {60.05$\pm0.3$}} & 46.27$\pm0.9$                        & {\color[HTML]{FF0000} 59.01$\pm1.0$}          & 38.83$\pm1.0$                        & 38.83$\pm1.0$                        \\\hline
        
           & AmazPantry & {\color[HTML]{FF0000} 67.83$\pm1.0$}          & 59.62$\pm0.8$                        & {\color[HTML]{FF0000} {67.99$\pm1.6$}} & {\color[HTML]{FF0000} 50.33$\pm1.2$} & {\color[HTML]{FF0000} 50.33$\pm1.2$} \\
AmazPantry & AmazDigiMu & {\color[HTML]{FF0000} {58.49$\pm0.8$}} & 51.48$\pm1.0$                        & {\color[HTML]{FF0000} 58.40$\pm0.5$}          & {\color[HTML]{FF0000} 42.71$\pm0.8$} & 42.71$\pm0.8$                        \\
           & AmazInstr  & {\color[HTML]{FF0000} 64.91$\pm0.5$}          & 54.92$\pm1.7$                        & {\color[HTML]{FF0000} {65.55$\pm1.0$}} & {\color[HTML]{FF0000} 43.31$\pm0.9$} & {\color[HTML]{FF0000} 43.31$\pm0.9$} \\ \hline
    
           & AmazInstr  & 69.52$\pm0.7$                                 & {\color[HTML]{FF0000} 63.06$\pm0.6$} & {\color[HTML]{FF0000} {70.73$\pm0.2$}} & {\color[HTML]{FF0000} 47.47$\pm1.0$} & 47.47$\pm1.0$                        \\
AmazInstr  & AmazDigiMu & 58.59$\pm0.8$                                 & 51.64$\pm0.4$                        & {\color[HTML]{FF0000} {58.93$\pm0.5$}} & 43.68$\pm0.7$                        & 43.68$\pm0.7$                        \\
           & AmazPantry & {\color[HTML]{FF0000} 64.95$\pm0.9$}          & 55.82$\pm0.6$                        & {\color[HTML]{FF0000} {65.58$\pm0.2$}} & {\color[HTML]{FF0000} 45.24$\pm0.8$} & 45.24$\pm0.8$             \\ \hline         
\end{tabular}
    \caption{Macro F1 and standard deviations of FRESH models with Top-$k$ explanations. {\color[HTML]{FF0000}{RED}} means \textsc{REGEX} outperforms the baseline.}
    \label{Fresh:full_results}
\end{table*}
\begin{table*}[!t]
    \centering
    \resizebox{\linewidth}{!}{
    \begin{tabular}{l|cccccc|cccccc}
    \hline
        \multirow{2}{*}{\textbf{Methods}} & \multicolumn{6}{c|}{\textbf{Normalized Sufficiency ($\uparrow$})} & \multicolumn{6}{c}{\textbf{Normalized Comprehensiveness ($\uparrow$})}    \\ \cline{2-7} \cline{8-13}
         & \textsc{Rand} & \textbf{$\boldsymbol \alpha \nabla \alpha$} &   \textbf{$\boldsymbol \alpha$} &    \textbf{DeepLift} &   \textbf{$\boldsymbol x \nabla x$} &          \textbf{IG}  & \textsc{Rand} & \textbf{$\boldsymbol \alpha \nabla \alpha$} &   \textbf{$\boldsymbol \alpha$} &    \textbf{DeepLift} &   \textbf{$\boldsymbol x \nabla x$} &          \textbf{IG} \\ \hline
       
        Saliency (Mean) & .32	& .52	& .32	& .48	& .44	& .45	& .25	& .48	& .53	& .42	& .40	& .38  \\ 
        InputXGrad (Mean) & .40	& .52	& .43	&.53	&.42	&.42	&.18	&.37	&.19	&.39	&.22	&.22  \\
         DeepLift (Mean) & .36	&.61	&.42	&.58	&.50	&.51	&.22	&.52	&\bf{.66}	&.49	&.37	&.37  \\ 
        IG (Mean) & .29	&.47	&.37	&.45	&.29	&.27	&.24	&.49	&.26	&.51	&.28	&.33  \\ \hline
        Saliency ($\ell_{2}$) & .32	&\bf{.70}	&.36	&.65	&\bf{.54}	&\bf{.54}	&.17	&.55	&.20	&.43	&.37	&.37  \\ 
        InputXGrad ($\ell_{2}$) & .34	&.58	&.38	&.54	&.43	&.43	&\bf{.29}	&.\bf{.58}	&.25	&.49	&.31	&.30  \\ 
        DeepLift ($\ell_{2}$) & .30	&.69	&.39	&.68	&.53	&.53	&.16	&.53	&.26	&.47	&.37	&.37  \\ \hline
        \approach{} & .30	&.68	&\bf{.48}	&\bf{.71}	&.49	&.49	&.22	&.56	&.41	&\bf{.52}	&\bf{.43}	&\bf{.43} \\ 
        \quad w/o robustness improvement & \bf{.38} & .54 & .43 & .42 & .42 & .42 & .12 & .33 & .18 & .17 & .20 & .20 \\ 
        \quad \quad w/o virtual adversarial training & .27 & 	.47	& .32	& .31	& .33	&.33	&.14	&.39	&.21	&.19	&.24	&.24 \\ 
        \quad \quad w/o input gradient regularization & .23	&.54	&.30	&.32	&.40	&.40	&.19	&.57	&.25	&.28	&.40	&.40 \\ 
        \quad w/o explanation guided training & .32 &	.66	& .40 &	.64	& .54	& .54	& .16	& .54	& .27	& .44	& .39	& .39  \\ \hline
    \end{tabular}}
    \caption{Full results of ablation study with different aggregation methods and  feature attribution methods in \S\ref{section:egt}.}
    \label{tab:full_ablation}
\end{table*}




\end{document}